\documentclass[11pt]{article}
\usepackage{PRIMEarxiv}
\usepackage[T1]{fontenc}
\usepackage[utf8]{inputenc}
\usepackage{graphicx}
\usepackage{booktabs}
\usepackage{tabularx}
\usepackage{multirow}
\usepackage{adjustbox}
\usepackage{subcaption}
\usepackage{caption}
\usepackage{svg}
\usepackage{amsmath}
\usepackage{xcolor}
\usepackage{hyperref}
\hypersetup{
  colorlinks=true,
  linkcolor=blue,
  citecolor=blue,
  urlcolor=blue
}
\newcommand{\orcid}[1]{\href{https://orcid.org/#1}{\includegraphics[width=10pt]{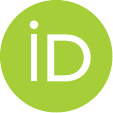}}}
\newcommand\blfootnote[1]{%
  \begingroup
  \renewcommand\thefootnote{}\footnote{#1}%
  \addtocounter{footnote}{-1}%
  \endgroup
}

\begin{document}
\title{End-to-End Chatbot Evaluation with Adaptive Reasoning and Uncertainty Filtering}
%
%
\author{
Nhi Dang \orcid{0009-0009-0266-9246} \and
Tung Le \orcid{0000-0002-9900-7047} \and
Huy Tien Nguyen \orcid{0000-0002-9948-1048} 
\\[0.8em]
Faculty of Information Technology, University of Science, Ho Chi Minh City, Vietnam\\
Vietnam National University, Ho Chi Minh City, Vietnam\\[0.5em]
\small\texttt{23C11039@student.hcmus.edu.vn, \{lttung, ntienhuy\}@fit.hcmus.edu.vn}\\
}
%

%

%
\maketitle              
\blfootnote{Corresponding author: Huy Tien Nguyen (\href{ntienhuy@fit.hcmus.edu.vn}{ntienhuy@fit.hcmus.edu.vn}).}
\begin{abstract}
Large language models (LLMs) combined with retrieval augmented generation have enabled the deployment of domain-specific chatbots, but these systems remain prone to generating unsupported or incorrect answers. Reliable evaluation is therefore critical, yet manual review is costly and existing frameworks often depend on curated test sets and static metrics, limiting scalability. We propose an end-to-end automatic evaluator designed to substantially reduce human effort. Our system generates Q\&A pairs directly from the underlying knowledge base, uses LLMs to judge chatbot responses against reference answers, and applies confidence-based filtering to highlight uncertain cases. Applied to a Vietnamese news dataset, the evaluator achieves high agreement with human judgments while significantly lowering review overhead. The framework is modular and language-agnostic, making it readily adaptable to diverse domains. This work introduces a practical, scalable solution for evaluating chatbots with minimal reliance on manual intervention.

\end{abstract}
\section{Introduction}
Large language models (LLMs) have seen explosive growth in both capability and adoption over the past two years, powering applications across diverse domains such as education, finance, healthcare, and e-commerce~\cite{raza2025industrial}. A particularly successful approach is Retrieval Augmented Generation (RAG), where LLMs are combined with external knowledge retrieval to produce up-to-date, context-aware responses. This approach underlies many specialized question-answering chatbots deployed by enterprises and research organizations~\cite{10.1145/3637528.3671470}.

Despite these advances, LLM-based chatbots still suffer from a fundamental limitation: hallucination. These models can generate responses that are factually inaccurate, misleading, or entirely unsupported by any retrieved or underlying source. Compounding this issue, retrieval mechanisms may surface incomplete, outdated, or irrelevant documents, thereby degrading the quality and reliability of the model's outputs \cite{10.1145/3637528.3671470,ji2023survey}. Such shortcomings pose significant challenges for both end users and system developers, as unreliable answers not only diminish user trust but also risk serious downstream consequences—including the spread of misinformation, flawed decision-making, and unintended harm in sensitive applications.

To ensure quality, developers typically rely on manual created test sets and human annotation to evaluate chatbot performance. While accurate, this process is time-consuming and does not scale easily to new domains or rapidly changing content. Frameworks such as DeepEval \cite{Confident-Ai} and RAGAS \cite{es-etal-2024-ragas} have recently broadened their scope by supporting synthetic dataset generation and metric-based evaluation (e.g., faithfulness, relevance, context recall). However, these toolkits typically require developers to orchestrate dataset creation, evaluation, and filtering as separate steps, and they emphasize numeric scoring rather than end-to-end automation. Furthermore, numeric metrics often yield ambiguous correctness values without clear decision boundaries, making it difficult to distinguish between fully correct, unsupported, or unanswered cases.

\begin{figure}[t]
  \centering
  \includegraphics[width=1\linewidth]{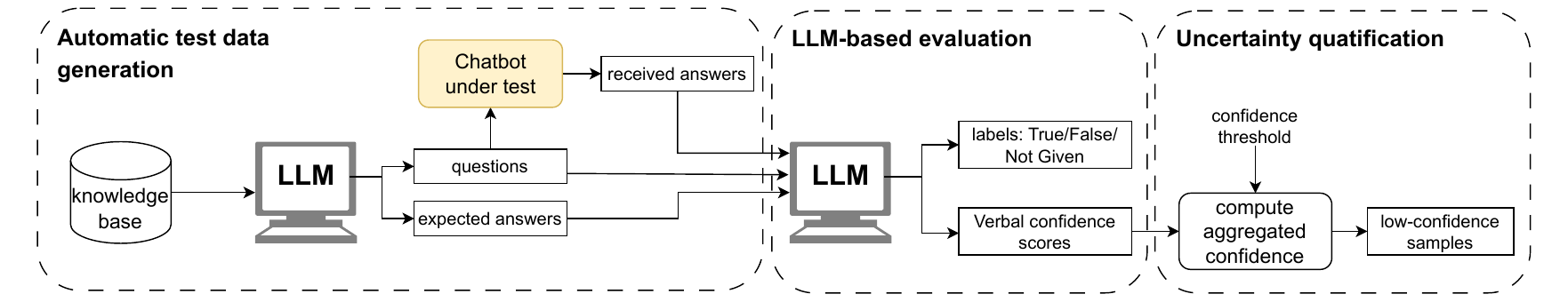}
  \caption{Overall system pipeline. The framework consists of three main components: (1) automatic test data generation, where LLMs generate questions and expected answers from an article database and query the target chatbot for responses; (2) LLM-based evaluation, where another LLM judges the correctness of responses and outputs both labels and verbal confidence scores; and (3) uncertainty quantification, which computes an aggregated confidence and filters out low-confidence samples for human review.}
  \label{fig:FullPipeline}
\end{figure}

In contrast, we present an automatic evaluation framework in which the chatbot itself is the only required input. Our system, shown in Figure \ref{fig:FullPipeline}, (i) generates domain-specific Q\&A pairs directly from the chatbot’s knowledge base, (ii) evaluates the chatbot’s responses using LLM-as-judge strategies and (iii) integrates step-wise confidence aggregation to filter low-trust cases for human review. To address the interpretability limitation of numeric scores, our evaluator produces categorical labels (TRUE, FALSE, and NOT GIVEN) that clearly distinguish factual errors from non-answers. This design transforms chatbot evaluation into an autonomous loop that produces both interpretable categorical judgments and confidence scores, thereby reducing human workload without sacrificing accuracy. Our key contributions are:
\begin{enumerate}
    \item \textbf{End-to-end automation:} a unified pipeline that requires no manually labeled test set, only the target chatbot and its underlying knowledge.
    \item \textbf{Interpretable LLM-as-judge evaluation:} generate labels with high reliability, accompanied by explanations that provide clearer diagnostics than score-only metrics.
    \item \textbf{Confidence-aware filtering:} aggregated per-step confidence to prioritize human review, reducing annotation cost while maintaining reliability.
    \item \textbf{Empirical validation: } experiments on a Vietnamese news dataset showing high agreement with human judgment and substantial savings in review effort.
\end{enumerate}

Together, these advances yield a scalable and practical framework for chatbot evaluation that can be applied across diverse domains and knowledge bases.

\section{Related work}

\subsection{Automatic test data generation with LLMs}
LLMs have been successfully used to generate synthetic Q\&A data from source texts, reducing the need for costly human annotation. For example, GPT-4 can be prompted to create Q\&A datasets in low-resource languages, producing questions with reasonable general knowledge \cite{putri-etal-2024-llm}. Another line of work focuses on converting documents into conversational Q\&A pairs. The \textit{Dialogizer} approach improved upon a prior "dialog inpainting" method by introducing a novel training framework to generate contextually relevant multi-turn Q\&A dialogs from textual sources \cite{hwang-etal-2023-dialogizer}. This system uses dual training objectives and a reranking step to yield higher-quality Q\&A pairs. Additionally, the \textit{SciQAG} framework fine-tunes an open LLM to distill Q\&A pairs from research papers, evaluating each pair for quality across multiple dimensions \cite{wan2024sciqag}. Collectively, these methods demonstrate that prompting or fine-tuning LLMs to automatically produce Q\&A test data from input documents is a viable strategy, yielding datasets comparable to or better than manually crafted resources.

\subsection{LLMs as automatic judges of model outputs}
Using powerful LLMs as automatic evaluators for chatbot responses and summaries has gained significant attention. Studies have shown that LLMs like GPT-4 acting as evaluators can closely approximate human judgment, achieving around 80\% correlation with human preference rankings \cite{zheng2023judging}. However, there are limitations to the reliability of the LLM-as-judge approach. For instance, LLM evaluators outperform standard automatic metrics but still fall short of fully replacing human judgments, particularly when distinguishing high-quality outputs \cite{shen-etal-2023-large}. The effectiveness of LLM evaluators also strongly depends on prompt design. Simply prompting an LLM to produce a numeric score is suboptimal; instead, requiring the model to provide explanations or rationales significantly improves alignment with human evaluations \cite{chiang-lee-2023-closer}. Prompting the LLM to explain its decisions consistently leads to better correlations with human judgments. Overall, while LLMs are effective zero-shot or few-shot evaluators, careful prompt engineering such as step-by-step reasoning is essential to achieving human-level reliability.

\subsection{Uncertainty quantification and Filtering in LLM evaluations}

Recent work has focused on uncertainty quantification techniques to determine when an LLM’s evaluation should be trusted or referred to human reviewers. One effective method involves prompting the model to output a confidence score alongside its answer or evaluation. For example, fine-tuning an LLM on Q\&A tasks to produce calibrated confidence scores enables selective abstention, significantly enhancing evaluation reliability \cite{chen-etal-2023-adaptation}. Additionally, prompting the LLM to justify its judgments through multi-step reasoning helps produce more consistent and human-aligned evaluations \cite{chiang-lee-2023-closer}. Another approach involves framing generation tasks as multiple-choice questions with an explicit option to express uncertainty, leveraging the model’s token-level probabilities to reflect confidence accurately \cite{ren2023self}. An effective practical implementation of uncertainty filtering uses the LLM’s confidence scores to automatically remove low-quality Q\&A pairs, thereby improving the overall dataset quality \cite{wan2024sciqag}. In essence, contemporary research emphasizes the importance of uncertainty-aware techniques, such as confidence scoring and multi-step reasoning, to improve the reliability of automated evaluation pipelines by filtering uncertain cases before finalizing judgments.

\subsection{Evaluation Frameworks}
A number of evaluation frameworks have recently emerged to support the assessment of LLM-based and retrieval-augmented systems, with DeepEval and RAGAS being among the most widely adopted. DeepEval \cite{Confident-Ai} is an open-source framework that positions itself as a testing library for large language models, providing developers with a wide range of built-in metrics such as faithfulness, answer relevancy, hallucination detection, and contextual recall. It has recently been extended with the ability to generate synthetic datasets, reducing the reliance on fully human-labeled resources. Despite these strengths, DeepEval is primarily a modular toolkit: dataset creation, metric computation, and downstream filtering remain separate processes, and the results are presented mainly as numeric scores. This makes it highly flexible, but less focused on providing end-to-end autonomous evaluation loops or interpretable correctness judgments.

RAGAS \cite{es-etal-2024-ragas} takes a complementary perspective by targeting RAG pipelines specifically. It provides metrics for both the retrieval and generation stages, including measures of context precision, context recall, faithfulness, and answer relevancy. Like DeepEval, RAGAS has also incorporated mechanisms for synthesizing evaluation data, enabling the creation of test sets from raw text without extensive human annotation. Its metric-driven approach is particularly well suited for diagnosing retrieval quality alongside generative accuracy. However, RAGAS remains centered on quantitative scores, and does not natively provide categorical judgments, reasoning chains, or confidence-based filtering strategies that can selectively escalate uncertain cases for human review.

\section{Proposed method}

In this section we present the three core components of our system: \textbf{(i) automatic test data generation}, which creates question–answer pairs for evaluation; \textbf{(ii) LLM-based evaluation}, which determines the correctness of chatbot responses; and \textbf{(iii) uncertainty quantification}, which filters out low-confidence cases for manual review.

\subsection{Automatic test data generation}

\begin{figure}[t]
  \centering
  \includegraphics[width=0.8\linewidth]{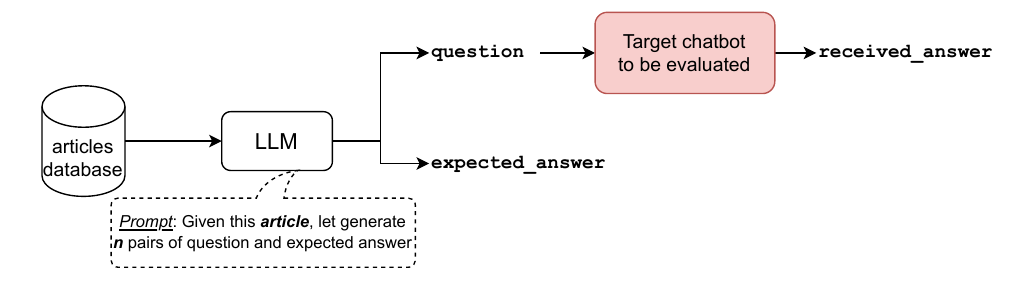}
  \caption{Automatic Test Data Generation Pipeline.
Given an article from the input database, an LLM is prompted to generate n pairs of question and expected answer. The questions are used to query the target chatbot, while the expected answers serve as ground truth for later evaluation.}
  \label{fig:DatesetGeneration}
\end{figure}

To create evaluation-ready Q\&A pairs without relying on manually labeled data, we designed a simple pipeline that uses LLMs to generate both questions and reference answers from real-world articles. The input to this process is a collection of Vietnamese news articles, each stored with its title, full content, publication date, and other metadata in a central database.

For each article, the system sends its content to an LLM along with a fixed prompt instructing the model to generate a number of question–answer pairs. The prompt encourages the model to produce questions that are either factual or based on simple inference, and to provide concise answers that are directly grounded in the article. The model returns a list of question and expected-answer pairs, which are extracted and stored for later use.

As illustrated in Figure \ref{fig:DatesetGeneration}, each generated question is sent to the chatbot under evaluation, which returns a corresponding answer. The LLM-generated expected answer is retained as a reference. These two outputs - the chatbot’s answer and the expected answer - form the basis for subsequent comparison and evaluation. This design ensures that all test cases are automatically generated, contextually relevant, and traceable to a common information source.

\subsection{LLM-Based automatic evaluation}
To assess the quality of chatbot-generated answers, we employ LLMs as automatic evaluators. The core idea is to present each evaluation instance - comprising the original question, the expected answer, and the chatbot’s received answer - to a judging LLM, which then outputs a label: TRUE, FALSE, or NOT GIVEN (meaning chatbot refuses to answer the question). This three levels labeling scheme allows the evaluator to separate incorrect answers (FALSE) from non-answers (NOT GIVEN), enabling the framework to differentiate between factual hallucinations and retrieval or abstention cases. Such separation is crucial for practical debugging of RAG-based chatbots, where the underlying cause of an error determines whether to adjust the generation model or to expand the knowledge base.

We implement three methods of increasing complexity for this judgment process: Single Prompt, Sequential Decision, and Adaptive K-step Reasoning.

\subsubsection{Single Prompt}
\begin{figure}[b]
  \centering
  \includegraphics[width=1\linewidth]{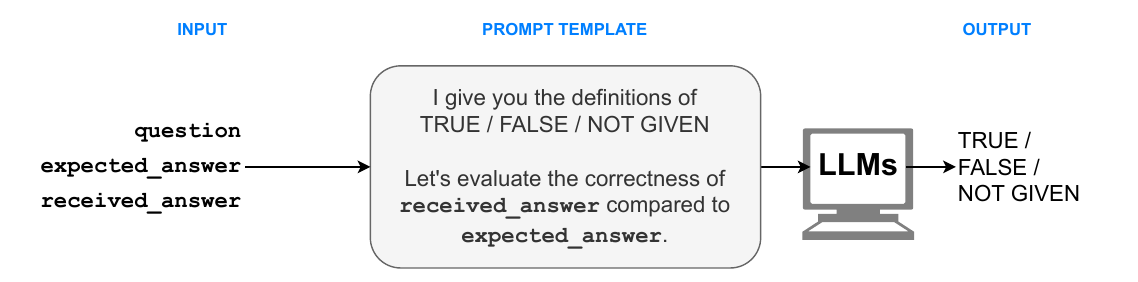}
  \caption{Single Prompt Evaluation Method.
The LLM receives a question, expected answer, and chatbot-generated answer, then directly returns a label (TRUE, FALSE, or NOT GIVEN) based on a predefined prompt template. }
  \label{fig:SinglePrompt}
\end{figure}

The simplest method involves a direct one-shot prompt to the LLM (see Figure \ref{fig:SinglePrompt}). The prompt includes a short definition of each possible label (TRUE, FALSE, NOT GIVEN) and then asks the LLM to evaluate whether the received answer is correct with respect to the expected answer. The model is expected to output one of the three labels directly.

This approach requires only a single inference call per sample, making it highly efficient. However, it does not allow the model to justify its decision or signal uncertainty, which limits its reliability in ambiguous or borderline cases.

\subsubsection{Sequential Decision}
\begin{figure}[t]
  \centering
  \includegraphics[width=1\linewidth]{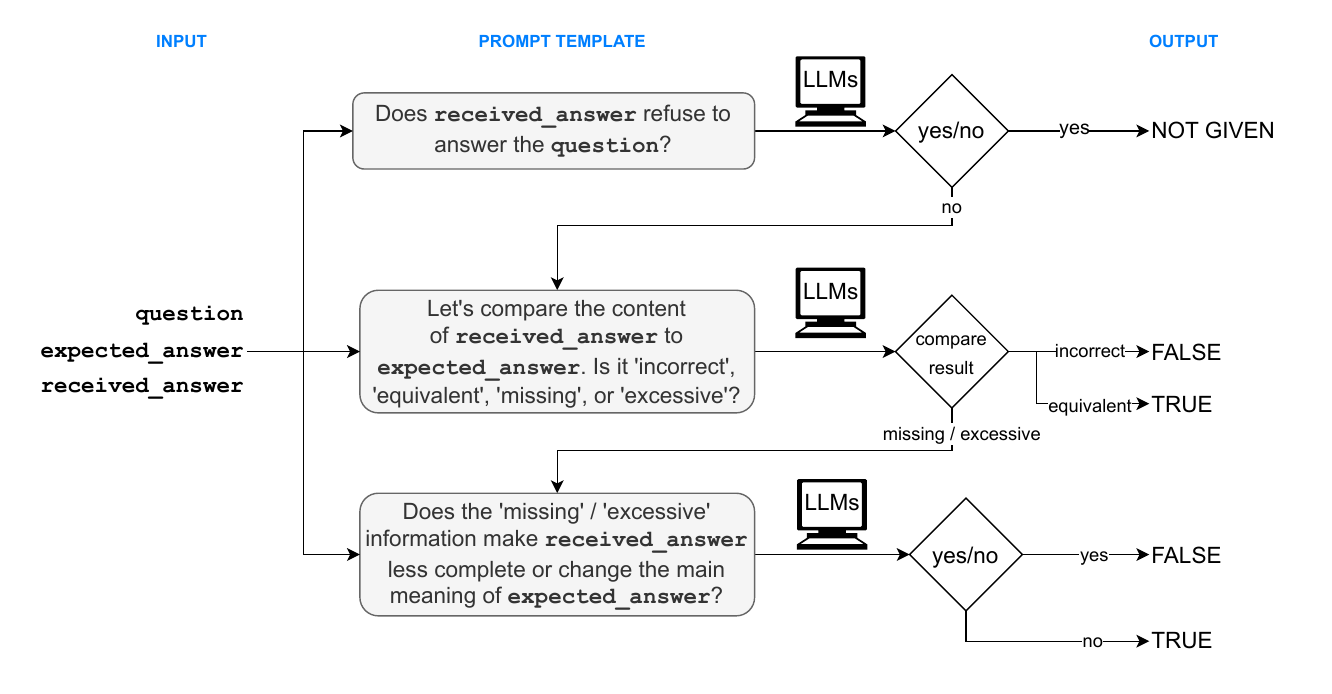}
  \caption{Sequential Decision Evaluation Method.
The LLM evaluates the received answer in a step-by-step manner. It first checks whether the answer refuses to respond, then compares it with the expected answer to classify the content as incorrect, equivalent, missing, or excessive. If additional or missing information is detected, the model decides whether it changes the core meaning. The final output is a label: TRUE, FALSE, or NOT GIVEN}
  \label{fig:SequentialSelection}
\end{figure}

The second method improves evaluation quality by breaking the judgment into structured steps (see Figure \ref{fig:SequentialSelection}). First, the model is asked whether the received answer avoids answering the question; if so, the label is NOT GIVEN. If not, the model compares the received answer to the expected answer and classifies it as equivalent, incorrect, missing, or excessive. Finally, if the answer includes missing or excessive information, the model assesses whether that affects the completeness or meaning of the original expected answer.

Each stage in this flow is prompted separately, giving the model a clearer framework to follow and allowing finer-grained reasoning. Although this approach still returns a final label only, the structured prompting helps mitigate overconfident errors from the model.

\subsubsection{Adaptive K-step Reasoning}
\begin{figure}[t]
  \centering
  \includegraphics[width=1\linewidth]{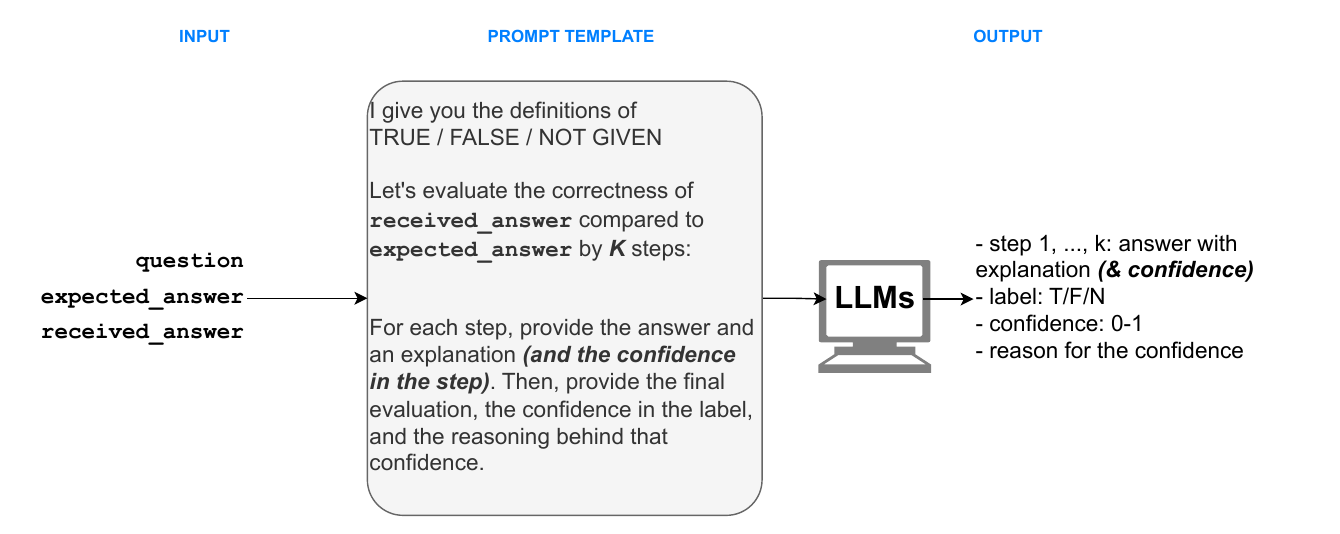}
  \caption{Adaptive K-step Reasoning Evaluation Method.
  The LLM is prompted to evaluate the received answer compared to the expected answer by reasoning through up to K self-defined steps. At each step, it provides a judgment, an explanation, and a confidence score. The final output includes the label (TRUE, FALSE, or NOT GIVEN), the overall confidence (0–1), and a rationale for the decision. }
  \label{fig:AdaptiveKstepReasoning}
\end{figure}

The most advanced method in our evaluation framework introduces a self-directed reasoning process controlled by a single hyperparameter K (see Figure \ref{fig:AdaptiveKstepReasoning}). Unlike the previous methods where the prompt specifies fixed sub-questions, here we allow the LLM to define and answer its own intermediate questions - up to a maximum of K reasoning steps.

The input includes the question, expected answer, and received answer, along with a prompt that instructs the model to carefully evaluate the correctness of the received answer in several steps. The model is free to choose how many steps to take (up to K) and what questions to ask itself at each step. After completing its reasoning process, the LLM outputs a final label (TRUE, FALSE, or NOT GIVEN), a confidence score between 0 and 1, and an explanation of the decision.

This method leverages the LLM’s internal reasoning capabilities to guide the evaluation process dynamically. It naturally captures uncertainty: samples requiring many steps or yielding low confidence can be flagged for human review, while confident, concise reasoning chains can be trusted as high-confidence labels. This flexibility makes it particularly well-suited for identifying borderline or ambiguous cases in an automated pipeline.

\subsection{Uncertainty quantification}
Building on our Adaptive K-step Reasoning framework, we adopt the stepwise confidence aggregation approach introduced by \cite{liu2024can}, which compute a single end-to-end confidence score by combining the per-step confidences produced during the model’s reasoning chain (see Figure \ref{fig:AdaptiveKstepReasoning}). At each reasoning step $i$ (where $i = 1, 2, \dots, m,\quad m \le K$), the LLM not only answers its self-posed sub-question but also emits a confidence value $c$ in the range $[0, 1]$. To capture the joint certainty of the entire chain, we compute the aggregated confidence

\begin{equation}
C \;=\;\prod_{i=1}^{m} c_i\label{eq:confidence}   
\end{equation}

This multiplicative aggregation treats each step’s confidence as an independent reliability estimate; a low value at any step proportionally lowers the overall score, reflecting the intuition that one weak link undermines the whole reasoning process.

Finally, we compare $C$ against a predefined threshold $\tau$. Instances whose aggregated confidence $C$ falls below $\tau$ are flagged for human review, while those at or above $\tau$ are accepted automatically as high-trust judgments. By filtering on this combined measure, our system focuses human effort only on the most uncertain or potentially error-prone cases, dramatically reducing manual workload without sacrificing evaluation accuracy.

\section{Experiment setup}

We conducted experiments on Vietnamese news to evaluate the effectiveness of the proposed evaluation strategy in assessing chatbot-generated responses and detecting unreliable outputs.

\textbf{Dataset.} The evaluation synthesized dataset consists of 300 question–answer pairs automatically generated from 50 Vietnamese news articles. For each article, an LLM was prompted to generate several factual or inference-style questions along with concise, source-grounded expected answers. To establish gold-standard labels, we asked three independent annotators to evaluate each pair by comparing the chatbot’s answer with the LLM-generated expected answer. Annotators assigned one of three labels: TRUE if the chatbot’s answer was correct and complete, FALSE if it was factually incorrect or contradicted the expected answer, and NOT GIVEN if the chatbot declined to answer or the response was irrelevant. In cases where the annotators disagreed, we applied majority voting to decide the final label. These annotated labels serve as the reference against which we assess the correctness of our system’s automatic judgments.

\textbf{Chatbot under test.} We evaluate a RAG-based chatbot that uses the same 50-article collection as its internal knowledge base. All articles were indexed using a sentence embedding–based retrieval method. For each input question, the chatbot retrieves the top 5 most relevant passages, which are then combined with the question and passed to gemini-1.5-flash-002 to generate a response. The resulting output serves as the chatbot’s received answer.

\textbf{Judge models.} For evaluation, we use six LLMs as automatic judges: three from the GPT family (gpt-4o-2024-08-06, gpt-4o-mini-2024-07-18, o1-mini-2024-09-12) and three from the Gemini family (gemini-1.5-flash-002, gemini-1.5-pro-002, gemini-2.0-flash-001). All judge models are queried with temperature set to 0.0 for consistency and reproducibility.

\textbf{Evaluation methods}.
We apply the three evaluation strategies introduced in Section 3: Single Prompt, Sequential Decision, Adaptive K-step Reasoning (with $K=3, 5, 7$). For the Sequential and Adaptive methods, we use the same confidence aggregation scheme and filtering threshold $\tau$ as described earlier to flag uncertain cases.

\textbf{Metrics}. We report three main evaluation metrics:

\begin{itemize}
    \item Accuracy: the proportion of model-predicted labels that match the gold human labels.
    \item False-label detection rate: the percentage of incorrectly labeled samples that are successfully flagged for review by the confidence threshold $\tau$.
    \item Human-review rate: the overall percentage of samples that are flagged as low-confidence  and thus require additional human verification.
\end{itemize}

This setup\footnote{Resources (dataset and source code) are released at https://github.com/nhidang912/ChatbotEvaluation} allows us to assess not only how accurate each evaluation method is but also how effectively it identifies unreliable chatbot responses that should be reviewed by humans.

\section{Evaluation and Analysis}

\subsection{Accuracy Comparison}

\begin{table}[t]
\centering
\caption{Accuracy by label for each model and evaluation method.}
\label{tab:evaluation_results}
\begin{adjustbox}{width=\textwidth}
\begin{tabular}{l|cccc|cccc|cccc}
\toprule
\multirow{2}{*}{Model} 
  & \multicolumn{4}{c|}{Single Prompt} 
  & \multicolumn{4}{c|}{Sequential Decision} 
  & \multicolumn{4}{c}{Adaptive K-step Reasoning (K=3)} \\
& T & F & N & Avg & T & F & N & Avg & T & F & N & Avg \\
\midrule
gemini-1.5-flash-002    & 0.94 & 0.74 & 0.47 & 0.72 & 0.97 & 0.55 & 0.97 & 0.83 & 0.93 & 0.84 & 0.67 & 0.81 \\
gemini-2.0-flash-001    & 0.98 & 0.65 & 0.83 & 0.82 & 0.96 & 0.65 & 0.93 & 0.85 & 0.92 & 0.81 & 0.73 & 0.82 \\
gemini-1.5-pro-002      & 0.95 & 0.74 & 0.90 & \textbf{0.86} & 0.97 & 0.55 & 0.93 & 0.82 & 0.92 & 0.84 & 0.87 & \textbf{0.87} \\
gpt-4o-mini-2024-07-18  & 0.95 & 0.77 & 0.83 & 0.85 & 0.93 & 0.65 & 0.93 & 0.84 & 0.79 & 1.00 & 0.80 & 0.86 \\
gpt-4o-2024-08-06       & 0.95 & 0.81 & 0.80 & 0.85 & 0.89 & 0.74 & 0.97 & 0.87 & 0.90 & 0.90 & 0.53 & 0.78 \\
o1-mini-2024-09-12      & 0.95 & 0.74 & 0.70 & 0.80 & 0.90 & 0.87 & 0.87 & \textbf{0.88} & 0.89 & 0.90 & 0.53 & 0.78 \\
\bottomrule
\end{tabular}
\end{adjustbox}
\end{table}

Because our gold-standard labels are imbalanced (239 TRUE, 30 FALSE, 31 NOT GIVEN), we use \emph{macro-average accuracy} to evaluate performance across all classes equally. The macro-average is computed by taking the average of the accuracies on each label type:
\begin{equation}
  \text{Macro‐Accuracy}
  \;=\;
  \frac{a_T + a_F + a_N}{3}\label{eq:accuracy}
\end{equation}

Table~\ref{tab:evaluation_results} shows the macro-average accuracies for six judge models under three evaluation methods: Single Prompt, Sequential Decision, and Adaptive K-step Reasoning (with \(K=3\)).

From the results, we observe that all methods perform well on the TRUE class - expected since it is the majority label in the dataset. However, this does not guarantee balanced evaluation across the board. In fact, the key difference between methods lies in their ability to handle the two much smaller classes: FALSE and NOT GIVEN.

Among the three methods, \textbf{Sequential Decision} offers the most consistent and stable performance. While not always achieving the absolute highest scores, it maintains strong accuracy across all labels for most models. This shows that breaking down the judgment process into guided substeps helps LLMs handle ambiguity and edge cases more reliably.

\textbf{Adaptive K-step Reasoning} comes very close, with the highest macro-average in some cases - particularly when using stronger models like GPT-4o-mini and Gemini-1.5-pro. Its strength comes from letting the model take control of the reasoning process. However, this flexibility can also introduce inconsistency for weaker models, especially on harder cases like NOT GIVEN. The approach works best when the underlying model is already strong at self-reflection.

\textbf{Single Prompt}, as the simplest method, is clearly limited. Although it does well on TRUE labels, its accuracy drops sharply on FALSE and NOT GIVEN, leading to the lowest macro-averages overall. This shows that a one-shot label prediction is not reliable enough in settings where subtle semantic differences matter - like deciding whether a generated answer is partially correct, irrelevant, or misleading.

Taken together, these results highlight the need for structured or self-reflective evaluation processes when aiming for reliable, class-balanced assessment. Strong judge models benefit more from self-directed reasoning (Adaptive), while structured prompting (Sequential) remains more stable across model tiers. Simple prompting may still be useful for quick, low-cost checks, but it cannot match the reliability of more advanced strategies.

\subsection{Filtering Uncertain Samples}
\begin{figure}[t]
  \centering
  \begin{subfigure}[t]{0.49\linewidth}
    \centering
    \includegraphics[width=\linewidth]{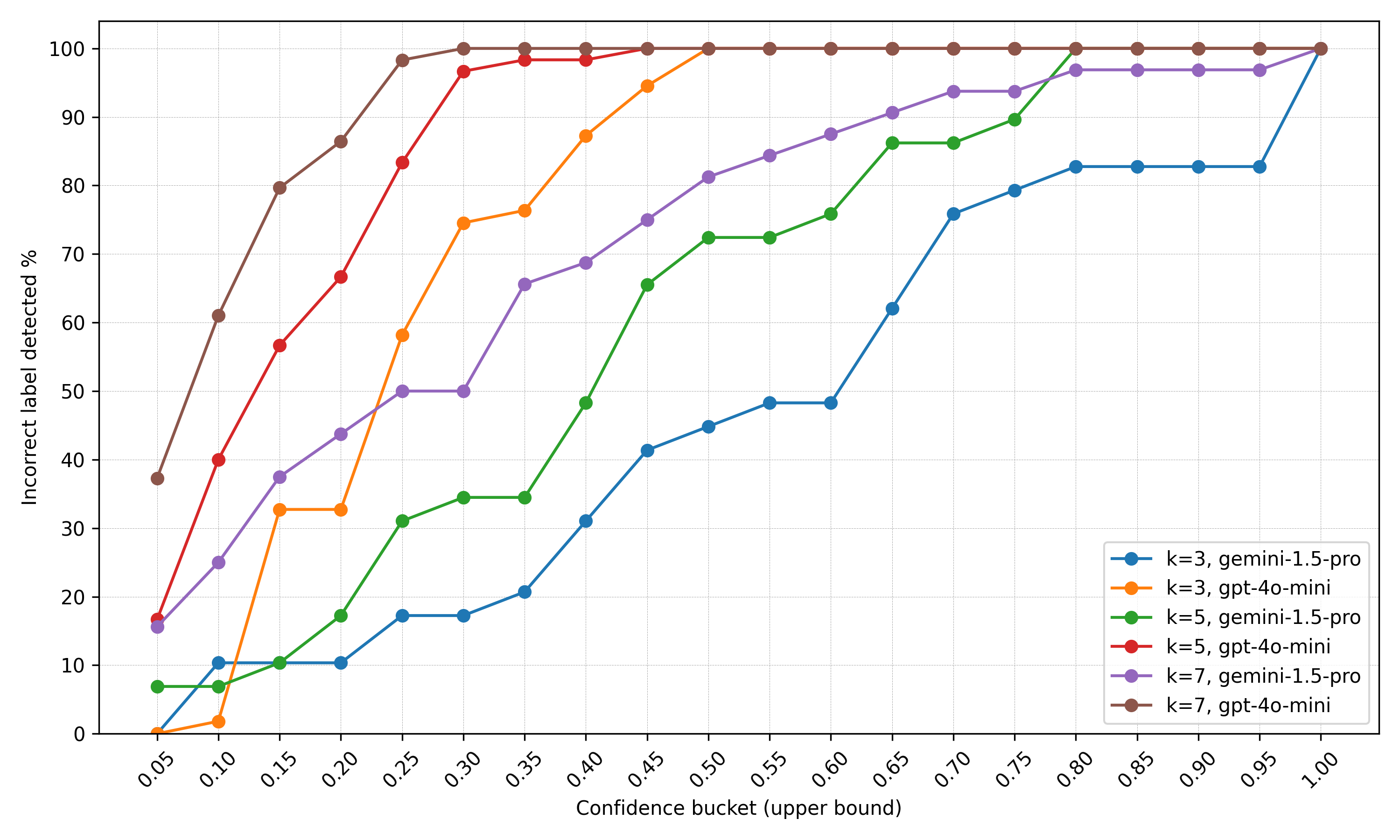}
    \caption{Error detection rate}
    \label{fig:IncorectLabelsDectectedPct}
  \end{subfigure}
  \hfill
  \begin{subfigure}[t]{0.49\linewidth}
    \centering
    \includegraphics[width=\linewidth]{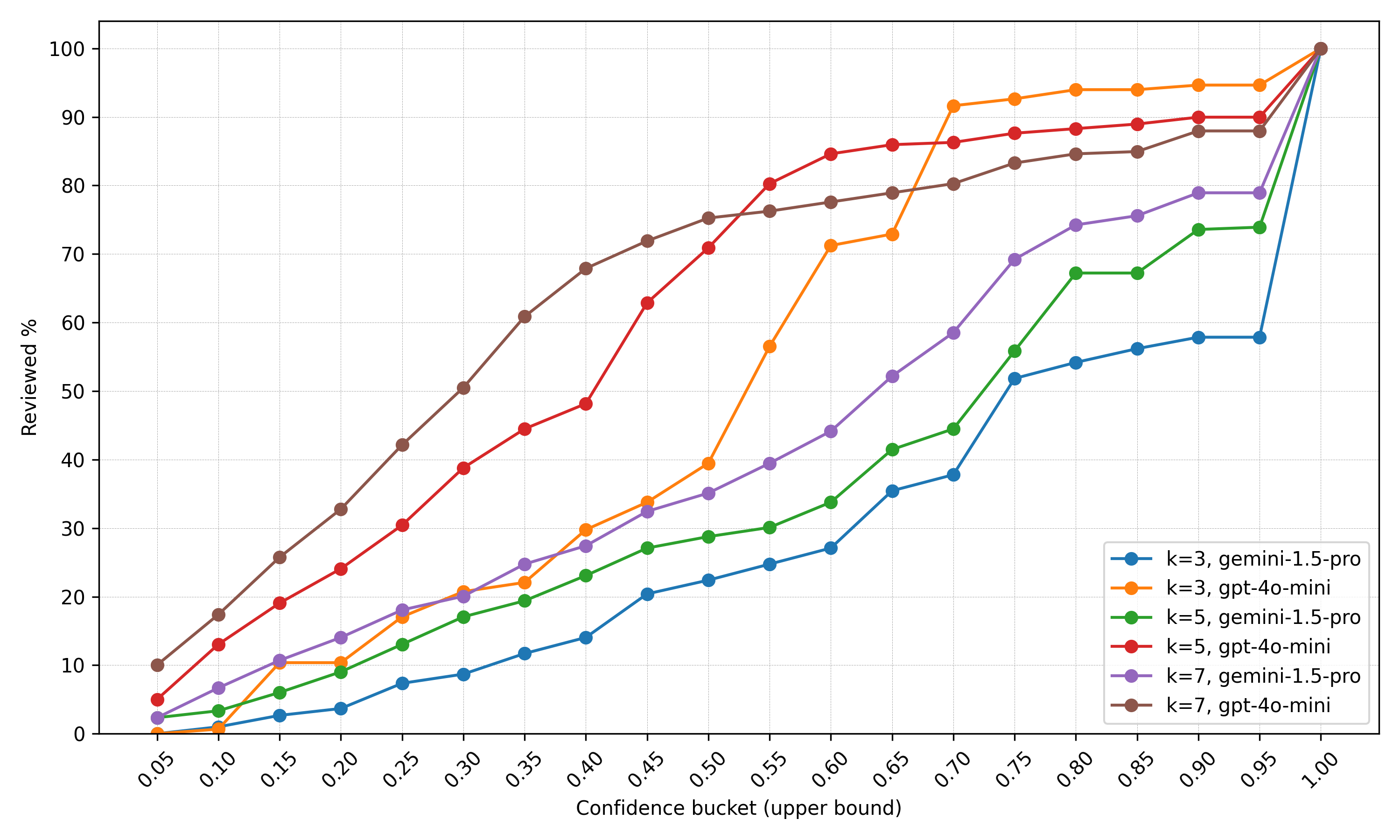}
    \caption{Human review percentage}
    \label{fig:HumanReviewPct}
  \end{subfigure}
  \caption{Performance of Adaptive K-step Reasoning: (a) detection rate of incorrectly labeled samples and (b) percentage of samples flagged for human review across confidence thresholds for different reasoning depths (K) and judge models.}
  \label{fig:confidence_analysis}
\end{figure}

Figures~\ref{fig:IncorectLabelsDectectedPct} and~\ref{fig:HumanReviewPct} illustrate how well the Adaptive K-step Reasoning method can help detect incorrect labels and reduce manual review, depending on the choice of threshold \(\tau\) and the number of reasoning steps \(K\). We focus on two judge models: gpt-4o-mini and gemini-1.5-pro.

The consistent alignment between confidence and labeling accuracy indicates that the model’s confidence values reflect genuine uncertainty about its own judgment rather than random variation, supporting their role as a practical trust indicator.

The chart in Figure~\ref{fig:IncorectLabelsDectectedPct} shows the percentage of incorrectly labeled samples that are successfully flagged for human review as \(\tau\) increases. As expected, a higher threshold allows the system to detect more errors, since more samples fall below the confidence limit and are sent for review. Across all values of \(K\), gpt-4o-mini consistently detects more errors at the same threshold compared to gemini-1.5-pro. For example, at \(\tau = 0.4\), gpt-4o-mini detects over 90\% of the incorrect labels at \(K=5\), while Gemini reaches only about 60\%.

The chart in Figure~\ref{fig:HumanReviewPct} shows the cost of achieving that detection: the percentage of total samples that would need to be manually reviewed at each threshold. As \(\tau\) increases, more samples fall below the threshold, so the review rate naturally rises. Again, gpt-4o-mini shows more favorable behavior. At \(\tau = 0.4\) and \(K=5\), it detects over 90\% of the mislabels while reviewing less than 30\% of the dataset. In contrast, gemini-1.5-pro at the same point needs to review around 45\% of the data for noticeably lower detection coverage.

This relationship between the two figures highlights an important insight: the threshold \(\tau\) acts as the control point to balance detection power and human-review effort. By adjusting \(\tau\), we decide how many samples to review, and in return, how many incorrect labels we can expect to catch. This gives developers the flexibility to configure the system based on real-world needs - whether they want maximum precision or reduced review workload.

Looking across different values of \(K\), we observe that increasing the chain length improves the ability to separate confident and uncertain cases - especially for gemini-1.5-pro. At \(K=3\), its detection curve rises slowly, and full coverage only happens near the maximum threshold. At \(K=5\), the curve shifts upward more sharply, and continues to improve at \(K=7\). However, this improvement comes with higher review cost. Unlike Gemini, gpt-4o-mini shows strong detection performance even at \(K=3\), but its review percentage also increases slightly as \(K\) increases. This suggests that gpt-4o-mini is able to reason confidently earlier, while deeper reasoning (higher \(K\)) tends to make the model more cautious and leads to more conservative (i.e., lower-confidence) responses that trigger review. 

In practical terms, this means that for high-performance models like gpt-4o-mini, even a shallow reasoning depth (\(K=3\)) already gives excellent results with minimal review. For models with more variability, such as gemini-1.5-pro, increasing \(K\) to 5 or 7 significantly boosts reliability and narrows the gap in performance.

Manual inspection of low-confidence samples shows that they often involve responses whose correctness is difficult to determine because of missing or additional details rather than clear factual errors. This means the uncertainty filter effectively highlights borderline cases where the answer’s reliability is uncertain, ensuring that human reviewers focus on responses where the uncertainty is genuine instead of random.

 For the question \textit{“What type of coins were found in the treasure?”}, the expected answer specifies 2,584 silver coins minted between 1066 and 1068, depicting William I and Harold II, while the chatbot answered only that the treasure “included silver coins.” During adaptive reasoning, the judge assigned relatively high confidence when verifying material type (0.7) but substantially lower confidence when evaluating missing details on quantity, minting period, and historical references (0.5 and 0.3), resulting in a low aggregated confidence score. Consequently, the sample was flagged for human review. In human annotation, some annotators considered the provided information sufficient at a minimal level, while others judged it incomplete, illustrating a genuinely ambiguous case appropriately captured by confidence-based filtering.

\section{Conclusion}
In this work, we have presented a fully integrated framework for automatically evaluating the reliability of RAG-powered chatbots. Our pipeline combines three key components: (1) automatic generation of domain-specific Q\&A pairs from chatbot's knowledge base, (2) an LLM-as-judge mechanism that compares chatbot answers against references using three methods: Single Prompt, Sequential Decision, and Adaptive K-step Reasoning - and (3) uncertainty quantification via  confidence-based filtering derived from multi-step reasoning outputs to flag low-trust responses for human review.

Our experiments on the 300-pair Vietnamese dataset generated by our system show that simple one-shot prompting suffices for many straightforward cases but struggles with nuanced errors, while Sequential Decision achieves the most balanced accuracy across TRUE, FALSE, and NOT GIVEN labels. Adaptive K-step Reasoning, when paired with appropriate thresholds, matches or exceeds Sequential Decision for strong models (e.g., gpt-4o-mini) and offers a flexible way to control the trade-off between automation and manual effort. By adopting categorical judgments instead of ambiguous numeric scores, our framework delivers interpretable and actionable diagnostics that clearly separate factual errors from non-answers. By tuning the confidence threshold, we can review as little as 30\% of the data while still catching over 90\% of the mislabels, reducing human workload by more than half without compromising evaluation quality.

Although our experiments focus on Vietnamese news articles, the proposed framework does not rely on language-specific features or domain-specific constraints and can be applied to other text-based knowledge sources. Our results across multiple judge models indicate that stronger models benefit more from adaptive multi-step reasoning, while weaker models may require shallower reasoning depths or higher human-review rates to maintain reliability. Moreover, the framework is most effective for fact-based, retrieval-grounded question answering, where expected answers can be traced to explicit sources. For more open-ended tasks, the higher degree of ambiguity may result in a large proportion of samples being flagged for human review, which can reduce the practical benefit of automation.

Despite its effectiveness, our framework still has room for improvement. The confidence threshold $\tau$ for filtering uncertain samples is manually selected, which limits robustness across models and domains. We envision an automatic threshold calibration mechanism, guided by a small validation set, that would allow the system to adapt seamlessly with little human intervention. Moreover, by enhancing the Sequential Decision method with lightweight clarification steps, we aim to reduce misclassifications in edge cases while preserving efficiency. These extensions will move our framework closer to a fully adaptive and generalizable evaluation pipeline.

%
%
%
\bibliographystyle{splncs04}
\bibliography{mybibliography}
\end{document}